\documentclass[journal,twocolumn]{IEEEtran}
\usepackage{amsfonts}
\usepackage{times}
\usepackage{graphicx}
\usepackage{latexsym}
\usepackage{dsfont}
\usepackage{amssymb}
\usepackage{amsmath}
\usepackage{cite}
\usepackage{verbatim}
\usepackage{subfigure}

\newcommand{\figref}[1]{{Fig.}~\ref{#1}}

\def\bb0{{\mathbb{0}}}

\def\bb{{\mathbf{b}}}

\def\b0{{\mathbf{0}}}

\def\sf0{{\mathsf{0}}}

\usepackage{epstopdf}
\usepackage{enumerate}
\usepackage{algorithmicx}
\usepackage{algorithm}
\usepackage{amsmath}
\usepackage[noend]{algpseudocode}
\usepackage{float}
\usepackage{hyperref}
\usepackage{color}
\usepackage{makeidx}
\usepackage{bbm}
\usepackage{graphicx}
\usepackage{lipsum}
\usepackage{subfigure}
\usepackage{tablefootnote}

\begin{document}

\title{ViWi: A Deep Learning Dataset Framework for Vision-Aided Wireless Communications}
\author{Muhammad Alrabeiah\textsuperscript{*}, Andrew Hredzak\textsuperscript{*}\thanks{*Authors contributed equally}\footnotemark{}, Zhenhao Liu, and Ahmed Alkhateeb\\ Arizona State University, Emails: \{malrabei, ahredzak, zliu294, alkhateeb\}@asu.edu}
\maketitle

\begin{abstract}
The growing role artificial intelligence and specifically machine learning is playing in shaping the future of wireless communications has opened up many new and intriguing research directions. This paper  motivates the research in the novel direction of  \textit{vision-aided wireless communications}, which aims at leveraging visual sensory information in tackling wireless communication problems. Like any new research direction driven by machine learning, obtaining a development dataset poses the first and most important challenge to vision-aided wireless communications. This paper addresses this issue by introducing the Vision-Wireless (ViWi) dataset framework. It is developed to be a parametric, systematic, and scalable data generation framework. It utilizes advanced 3D-modeling and ray-tracing softwares to generate high-fidelity synthetic  wireless and vision data samples for the same scenes. The result is a framework that does not only offer a way to generate training and testing datasets but helps provide a common ground on which the quality of different machine learning-powered solutions could be assessed.
\end{abstract}


\section{Introduction} \label{sec:Intro}


\textbf{Can we use vision to help wireless communication?}  There are several reasons that motivate asking this question. First, future wireless communication devices and base stations will likely employ large numbers of antennas (at sub-6GHz or mmWave band ) to satisfy the high data rate requirements \cite{Boccardi2014,Rappaport2013a}. These  large-scale  multiple-input multiple-output (MIMO) transceivers, however, are subject to critical challenges such as the requirement of large channel/beam training overhead  and the sensitivity of mmWave links to blockages \cite{Bjoernson2016,Alkhateeb2018,Rappaport2013a}. Interestingly, most of these devices that employ large-antenna arrays will most likely have other sensors, such as RGB cameras, depth cameras, or LiDAR sensors. This is the case, for example, in vehicles, 5G phones, AR/VR, intersection nodes of self-driving cars, and probably  base stations in the near future. It is therefore natural to ask whether this vision data (generated for example by  cameras) can help overcome the non-trivial wireless communication challenges, such as mmWave beams and blockage prediction, massive MIMO channel subspace prediction, hand-over prediction, and proactive network management among others. This is further motivated by the recent advances in deep learning and computer vision that can extract high-level semantics from complex visual scenes, and the increasing interest of leveraging machine/deep learning tools in wireless communication problems \cite{LIS, DLChMapping, OShea2018, Alkhateeb2018a,Li2019, Samuel2017}.

\textbf{The need for a dataset:}    To enable leveraging deep learning and computer vision  for the proposed \textit{vision-aided wireless communication}  research, it is crucial to have sufficiently large and suitable datasets.  These datasets will allow the researchers to (i) develop deep-learning/computer vision algorithms and evaluate their performance, (ii) reproduce the results of the other papers, (iii) set benchmarks for the various vision-aided wireless communication problems, and (iv) compare the different proposed solutions based on common data. Next, we describe the main requirements in such a dataset to be useful for  the vision-aided wireless communication research. 
\begin{itemize}	
	\item \textbf{Co-existing visual and wireless data:} Since the objective is to use visual data, captured for example by cameras or LiDAR sensors, to help the wireless communication systems that operate in the same device or environment, the visual data (such as the RGB and depth images as well as point cloud data) and the wireless data (such as the communication and radar channels) need to be collected from the same environment. 
	\item \textbf{Accuracy:} The methodology of collecting the visual and wireless data should ensure the accuracy of this data.  
	\item \textbf{Scalability:} The dataset collection process should be scalable to many scenarios and sizes to be able to efficiently address several use cases.
	\item \textbf{Parameterized dataset:} In wireless communication problems, it is normally important to evaluate the performance versus different system and channel parameters, such as the number of antennas and array geometry. Similarly, we expect that it will be desirable to study the vision-aided wireless communication algorithms for different visual-data parameters, such as the camera resolution, color space, depth, and point cloud perturbation. Therefore, the dataset that enables these research directions needs to be parameterized. 
\end{itemize}

There are several datasets that have been developed over the last decade for visual data alone  \cite{Geiger2012,Russakovsky2015}, or more recently for wireless data alone  \cite{DeepMIMODataset} or  wireless/LiDAR data \cite{Klautau}. To the best of our knowledge, however, there are no publicly available datasets  that provide co-existing visual and wireless data.

\

\begin{figure*}[t]
	\centering
	\includegraphics[width=.95\linewidth]{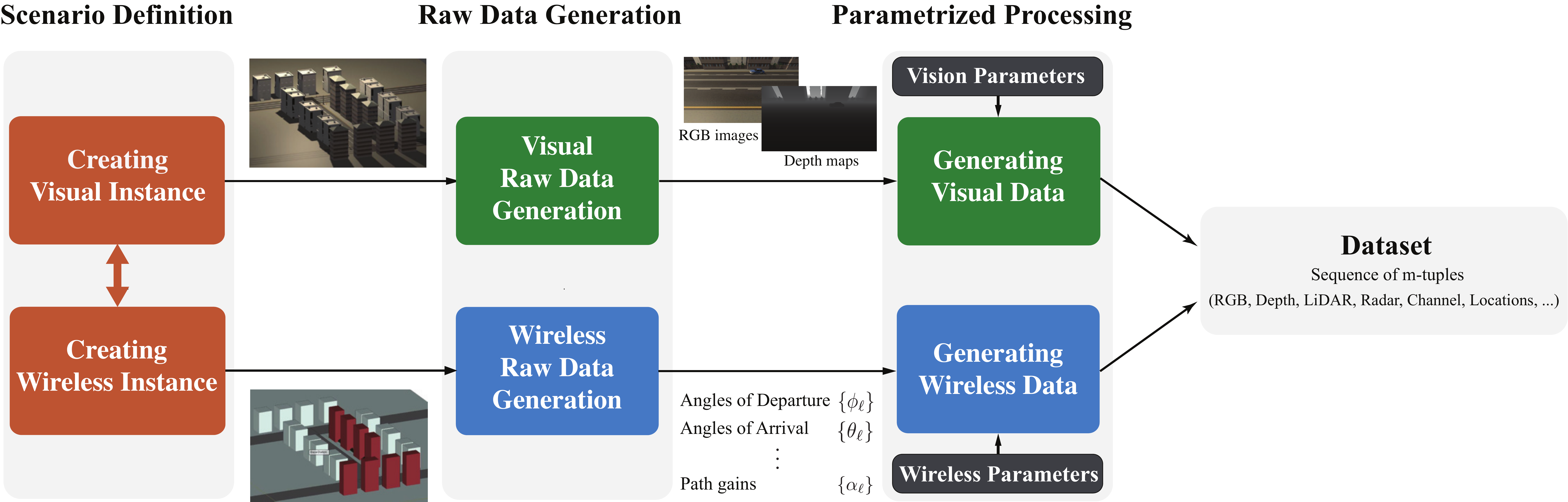}
	\caption{A block diagram for the overall ViWi dataset generation framework structure. It shows the main three stages of dataset generation, the elements of each stage, and some example outputs form each element.}
	\label{framework block-diagram}
\end{figure*}

\textbf{The ViWi dataset:} This paper presents the Vision-Wireless (ViWi)\footnote{The latest versions of the ViWi datasets  and codes can be found on the dataset website \cite{ViWi}.} framework that is designed to satisfy the mentioned requirements.  ViWi is a data-generating framework that does not only provide wireless data but combines it with visual data taken from the same scenes. This is achieved by utilizing advanced 3D modeling and ray-tracing simulators that generate high-fidelity synthetic  vision and wireless data. The main goal of creating the ViWi dataset framework is to encourage and facilitate research in vision-aided wireless communications, which utilizes the advances in computer vision, machine learning, and point cloud analysis to tackle the critical challenges in wireless communications. In the first release, we make four ViWi-generated datasets publicly available \cite{ViWi}. Each ViWi-dataset consists of 4-tuples of image, depth map, wireless channel, and user location. 

Before diving into the details, here is how this paper is structured. The next section, Section.\ref{overview}, provides an overview of ViWi and highlights its major components. Section \ref{Details} takes a deeper look into those components using two example scenarios, which results in the first two vision-wireless datasets. Section.\ref{applications} presents some possible applications for the framework. Finally, Section \ref{conclusion} concludes the paper.


\section{Framework Overview}  \label{overview}

The availability of a development dataset constitutes a major challenge to vision-aided wireless communications. As such, this work presents a novel framework for visual-wireless synthetic data generation. The choice of using synthetic data is mainly motivated by two factors: (i) its relatively-low cost and (ii) its scalability. Acquiring real-world visual and wireless data, like images and channels, requires two completely different equipment setups, and the data acquisition process itself is time consuming; the process entails building a physical scenario, placing the equipment, synchronizing the acquisition process, and collecting data over a lengthly period of time. All that translates into increased cost and difficult scalability when compared to generating synthetic data \cite{SynData-CV1}. These challenges have been acknowledged, albeit independently, by the computer vision and wireless communication communities. An increasing amount of work in both communities has been relying on synthetic data generated by 3D game engines and electromagnetic ray-tracing softwares, see for example \cite{Klautau} where synthetic data were used to build a dataset for joint Lidar and wireless systems, or \cite{DeepMIMODataset} where ray-tracing softwares were leveraged to construct a generic dataset generation framework for MIMO applications (other examples in \cite{SynData-CV1, Synthia, SIM4CV, LIS, DLChMapping}). Hence, advanced game engines and ray-tracing softwares are the backbone of the proposed Vision-Wireless (WiVi) dataset framework.

\begin{table*}[hbt]
	\centering
	\caption{A list of objects composing the visual and wireless instances of the scenario}
	\begin{tabular}{cccc}
		\hline\hline
		Object & Dimensions (Width, Length, Hight in meters) & Material & Note \\
		\hline\hline
		Model Building 1\textsuperscript{*} & $6.4 \times 9.7\times 21.2$ & Brick & Replaced with same-dimensions cube\\
		Model Building 2\textsuperscript{*} & $ 10\times 10\times 15$ & Concrete & Replaced with same-dimensions cube \\
		Street & $ 15\times 90 $ & Asphalt & \\
		Sidewalk & $ 3\times 90 $ & Concrete & \\
		Fence\textsuperscript{*} & $0.1\times 8.2\times 0.42 $ & -- & Removed \\
		Bush & $ 1\times 1\times 1 $ & Dense deciduous forest & \\
		Trafic Light\textsuperscript{*} & $ 0.9\times 5\times 6 $ & -- & Removed \\
		Fire Hydrant\textsuperscript{*} & $ 0.3\times 0.3\times 0.8 $ & -- & Removed \\
		Garbage Dumpster\textsuperscript{*} & $2\times 2.2\times  2.5$ & -- & Removed \\
		Car\textsuperscript{*} & $2\times 4.5\times 1.5$ &   -- & Replaced with user grid.\\
		Bus\textsuperscript{*} & $3\times 14\times 3$ & Metal & Replaced with same-dimensions cube \\
		\hline\hline
	\end{tabular}
	\label{ObjList}
\end{table*}

The dataset generation in the proposed framework goes through three main stages as shown in  \figref{framework block-diagram}. These stages are:
\begin{itemize}
	\item \textbf{Scenario definition:} Addressing a vision-wireless problem starts by describing the physical study environment where the problem is defined, which is referred to as \textit{scenario definition}. This description must identify two types of elements, visual and electromagnetic. The visual elements, e.g., buildings, curbs, streets, cars, trees, people...etc, are built and assembled using a game engine software. They all together form the visual instance of the scenario. The same scenario definition with its visual elements is constructed in a ray-tracing software. This software defines the electromagnetic characteristics of the scenario, like dielectric properties of different objects, creating the wireless instance of the scenario. See the left column in \figref{framework block-diagram}.
	\item \textbf{Raw-data generation:} The two scenario instances are processed by the game engine and the ray-tracing software to produce two sets of raw data. The first is a set of visual data, which are RGB images of the environment, accurate depth maps, and LiDAR point cloud, while the other set has wireless data such as angles of arrival/ departure and path gains of all the rays between the transmitters and  receivers. These two sets together define the scenario raw data. See the middle column in \figref{framework block-diagram}.
	\item \textbf{Parameterized processing:} This stage offers the choice of customizing the raw data using two sets of user-defined parameters. Both sets define how the visual and wireless raw data is processed to extract the final and, often, more realistic data samples. For visual raw data, this may include transforming images to different color spaces, lowering the resolution of images and depth maps, adding some artifacts to them, or distorting point cloud data. On the other hand, for the wireless data, this may include constructing wireless communication/radar channels and obtaining user locations. See the right column in  \figref{framework block-diagram}.
\end{itemize}

\textbf{What is interesting and unique about this three-stages framework is that every dataset is completely defined by its scenario name and the parameterization sets;} it is enough to provide the name of the scenario and the parameter sets to completely describe a certain dataset or generate (reproduce) it. This allows for fast and easy re-generation and makes the framework favorable for benchmarking.  



\section{ViWi: A Detailed Description} 
\label{Details}
With the aforementioned ViWi structure in mind, this section discusses the inner-workings of each stage using two example datasets\footnote{The first release of ViWi could, actually, be used to generate four different datasets using four different scenario raw data, but for the sake of clarity, only two of those four datasets are used as examples in the discussion.}, namely \textit{distributed-camera} and \textit{co-located-camera} datasets. Both of them are generated using the same example scenario as it is explained below.

\subsection{Scenarios Definition}\label{scenario}
The two example datasets generated with ViWi are for an outdoor scenario, which shows a car driving through a city street.  \figref{visual}-a depicts an areal view of the visual instance of this scenario. It is built and generated using the popular game engine Blender \texttrademark. This instance is composed of many elements that are found in real-world metropolitans, like buildings, bushes, sidewalks, cars,...etc. Table \ref{ObjList} lists the building blocks of the scenario and their dimensions. To animate the scenario, five trajectories are defined to represent possible car paths, each of which has one thousand equally spaced points that are 0.089 meters apart. The trajectories are also separated form each other with equal distance, which is 0.5 meters. This visual instance is used for both datasets but with some minor visual changes. More on that in Section \ref{Raw-Data}

To generate the wireless instance of the scenario, the visual instance is imported into the ray-tracing software of choice, which is, in this work, Wireless InSite\textsuperscript{ \textregistered }. Some of the objects in the visual instance have very fine visual details that may substantially slow down the ray-tracing simulation. In cases where hardware or software capabilities are limited, those objects could be either removed form the wireless instance of the scenario or replaced with objects of simpler geometry with no major impact on the simulation results.  \figref{visual}-b shows an example of how the visual instance of the example scenario is simplified for ray-tracing simulation. Once the complexity situation is settled, the wireless instance is completed by setting the dielectric properties of all its objects. Table \ref{ObjList} shows the material used for every object in the wireless instance and identifies which objects are removed or replaced. 



\subsection{Raw-Data Generation}\label{Raw-Data}
This stage prepares the instance for processing and generates visual and wireless raw data. The visual and wireless instances are both fitted with raw-data generators, like cameras, transmitters, and receivers, and their properties are set in preparation for data generation. Both instances are run separately to get the output data, which is in its initial form. The visual raw data of the generated two datasets consists of RGB images and depth maps whereas the wireless raw data is composed of the angles of departure, path gains, and channel impulse responses for every simulated ray from the transmitter to the receiver. 

\begin{figure}[t]
	\centering
	\subfigure[ ]{\includegraphics[width=0.84\linewidth]{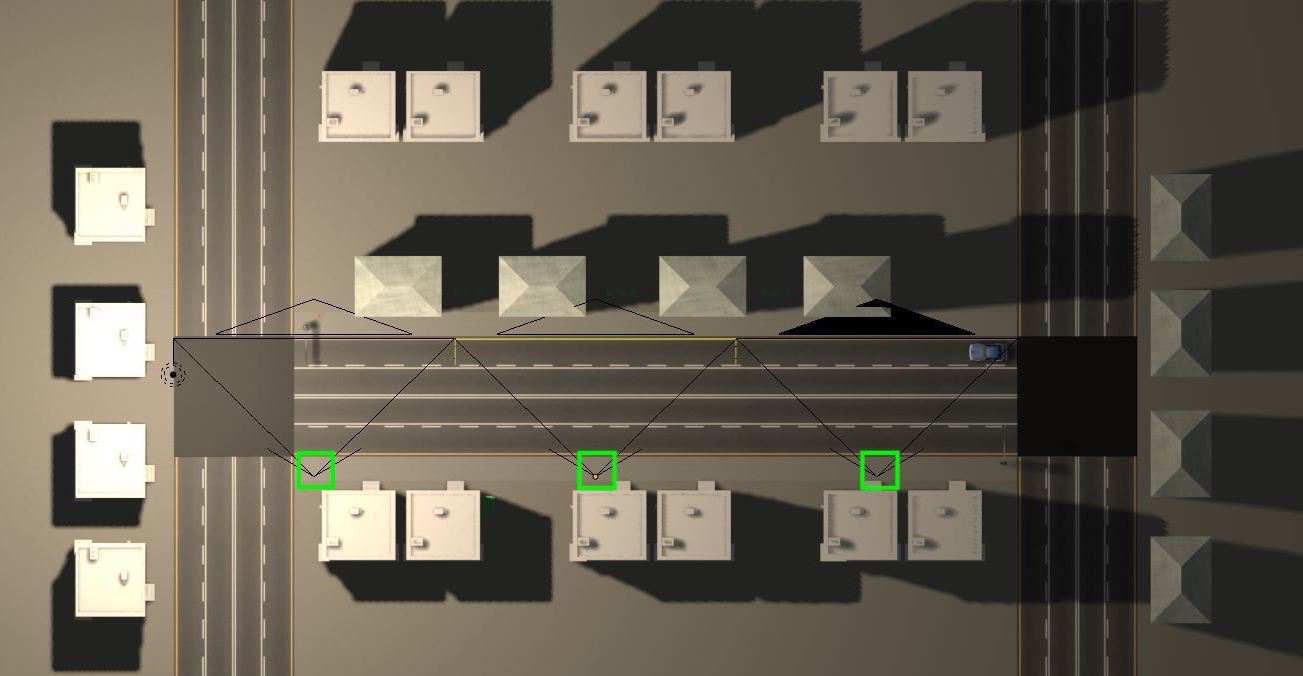}}
	\subfigure[ ]{\includegraphics[width=0.84\linewidth]{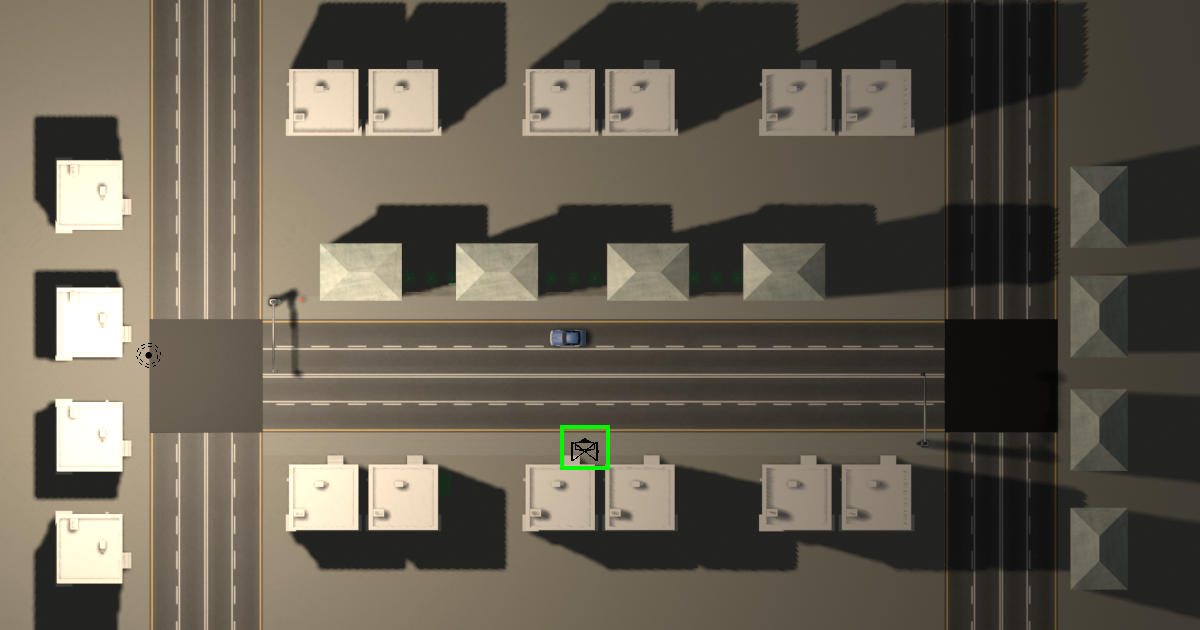}}
	\caption{Top view of the two example visual instances. Top image, (a), shows the locations of the distributed cameras where the bottom image, (b), shows the location of the centered cameras.}
	\label{cam_loc}
\end{figure}

The RGB images and depth maps in the two example datasets are produced from the same visual instance but using different visual data generators. For the first dataset, the visual instance is fitted with a total of 3 cameras (data generators) that are 5-meters high and 30-meters apart,  \figref{cam_loc}-a. Each camera has a field of view of 100 degrees. These properties are chosen so that the cameras cover the whole street with minimum field of view overlap. With these settings and using the defined car trajectories, the scenario is animated in Blender to generate the visual raw data of the fist example dataset, which will be henceforth referred to as the \textit{distributed-camera} dataset. For the second example dataset, three differently-oriented cameras with 75-, 110-, and 75-degree fields of view are placed half-way through the street and 5-meters above the ground,  \figref{cam_loc}-b. They are oriented in different directions such that they cover the whole street with the least possible overlap. Similar to the first example, the scenario is animated using the new generators and same car trajectories to produce the visual raw data of the second example dataset, which will be henceforth referred to as the \textit{co-located-camera} dataset. 

\begin{figure}
	\centering
	\subfigure[ ]{\includegraphics[width=0.7\linewidth]{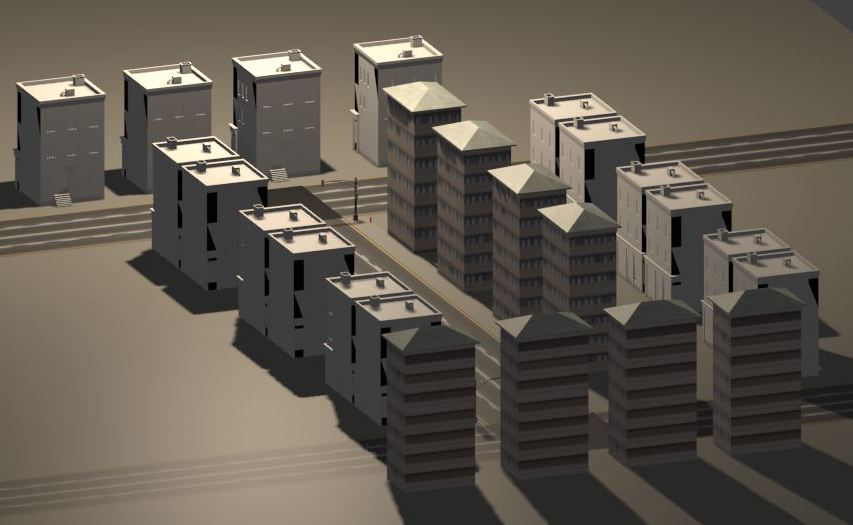}}
	\subfigure[ ]{\includegraphics[width=0.7\linewidth]{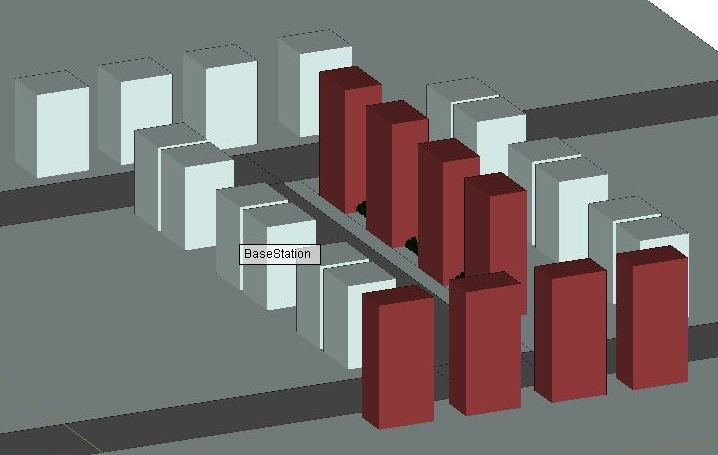}}
	\caption{Two images of the visual and wireless instances of the scenario. (a) is an aerial view of the visual instance while (b) is an aerial view of the wireless instance. They clearly shows the geometric changes between the two instances, e.g., no traffic lights and buildings have simpler geometry.}
	\label{visual}
\end{figure}

To generate wireless raw data of both example datasets, the wireless instance of the scenario,  \figref{visual}-b, is fitted with distributed data generators (transmitters and receivers) with similar properties. For both datasets, all transmitter and receiver antennas implement half-wave dipoles operating at a frequency of 60 GHz and with a sinusoid waveform. The first example has transmitter antennas, referred to as BaseStations (BSs), replacing the three distributed cameras, and a user grid of receiver antennas placed along each of the five pre-defined car trajectories. On the other hand, the second example has the three cameras replaced with one BS and uses the same five trajectories to define the receiver grid. Wireless InSite is used with both wireless instances to identify all possible rays going from every BS to every user in both examples, and produce two sets of wireless raw data, one for each dataset.



\subsection{Parametrized Processing}\label{Processing}
The raw data, whether visual or wireless, could be directly used as samples of a development dataset. However, studying real-world engineering problems and applications requires some form of control over the data acquisition process and the environmental settings. For instance, the quality of the camera feed or channel information could be subjects of interest in certain vision-aided communication problems. In such cases, the output raw data is in a primitive form that cannot be used to address any of those issues. Hence, raw data has to undergo another optional layer of processing to produce the final dataset, the last stage in the proposed framework. 

The last stage is a parametrized layer where the user defines how the raw data is processed. ViWi provides processing for the wireless and visual raw data independently in the form of a package of scripts. For the visual raw data, the scripts offer control over a set of parameters that chooses the scenario of interest and applies some filtering and transformation operations on the images or depth maps. Examples of such operations are image blurring filters, noise-corruption processes, resolution control, and color-space transformation. On the other hand, for the wireless raw data, other scripts define another set of control parameters. It includes specifying the scenario of interest, number of active BSs, number of antennas across x-, y-, and z-axes, and antenna spacing to name a few, the full list of wireless parameters and their definitions are the same as those of the DeepMIMO dataset \cite{DeepMIMODataset}. By setting those parameters, a user can produce a task-specific set of wireless data samples such as complex-valued channels and user locations.


\section{How to Use ViWi?}\label{Usage}
The first release of ViWi provides  four sets of raw data and a dataset-generating package \cite{ViWi}. Each set has the visual and wireless raw data required to generate the final dataset. This release provides only the scripts that parameterize the wireless raw data. Visual raw data do not undergo any processing, and, therefore, they are directly included in the final dataset. However, they are provided in popular data format, i.e., JPEG for RGB images and MAT data\footnote{MATLAB\textsuperscript{ \textregistered } native data structure, which could be easily read using other scripting languages like Python or R. The reason behind this choice is the popularity of MAT format compared to the original OpenEXR format used for depth maps.} for depth maps, so they could be easily processed by the user.

ViWi provides visual raw data in sub-directories enclosed in a main compressed directory ready for download. The images and depth maps are stored into two separate sub-directories. All raw images and depth maps have 720p HD resolution, i.e., $720$ pixels $\times$ $1280$ pixels, and are, as stated above, stored in JPEG and MAT format. Every image has a corresponding depth map, so they both have the same name in both subdirectories. More information on the naming system could be found in the "README.txt" file in the main compressed directory. Generating visual data samples of a dataset only requires unpacking (unzipping) the compressed directory.

The main compressed directory also contains a third sub-directory of MAT data files. Every BS (transmitter) in the wireless instance of the scenario contributes three MAT data files: (i) angles of departure file, (ii) complex impulse response file, and (iii) path gains file. This means there are 9 MAT files in the sub-directory of the distributed-cameras dataset and three files in that of the co-located-cameras dataset. To generate wireless data samples, the wireless raw data needs to be unpacked and processed using the ViWi script, which is provided separately. More technical details on how to generate the output wireless data and their structure could be found in the ``README.txt'' file enclosed with the script package.

\section{Possible Applications}
\label{applications}
Vision-aided wireless communications is a relatively new direction of research with a lot of potential.  With the ViWi framework, it is now possible to investigate more problems and benchmark more computer-vision-powered solutions. The following three subsections provide a rough categorization of the problems that could benefit from ViWi.
 
\subsection{Camera-Aided Beam Prediction}
Beam-prediction is a well-known problem in mmWave communications. Typical approaches to tackling this problem usually involve a form of beam-training with a fixed beam-forming codebook, which is usually time consumming. Some solutions have recently been proposed to utilize machine learning \cite{Alkhateeb2018} and reduce that training burden. However, all those solutions use exclusively wireless data to do beam prediction. The introduction of visual sensory data could make an interesting addition to the problem, for it provides a method to understand or analyze the surrounding environment of the transmitter and receiver. The two datasets produced with ViWi could be easily used to study such problem; both contain RGB images, depth maps, and channels for every user position.

\subsection{Blockage Prediction}
This is one of the most elusive problems not only in mmWaves but in wireless communications in general. It requires a strong sense of the surroundings and its dynamics as well as an intelligent analysis and prediction algorithm. The use of machine learning for predicting blockages has been investigated in \cite{Alkhateeb2018a,Alrabeiah2019a}, and the results are overall promising. The colocated-cameras dataset in ViWi provides an interesting scenario where more advanced solutions could be studied; along with wireless and depth data, it provides RGB images and user spatial locations. 


\section{Acknowledgment}
The authors thank Mr. Tarun Chawla and Remcom  for supporting and encouraging this work. The authors also thank Prof. Aldebaro Klautau from Federal University of Para for suggesting Blender to generate the synthetic images.  

\section{Conclusion}
For the interesting premise and massive potential of vision-aided wireless communications, this paper introduces the Vision-Wireless (ViWi) dataset generation framework. ViWi facilitates the research in this direction by offering a method for unified and modular generation of development datasets and for benchmarking different solutions. The current version of ViWi offers four datasets for four outdoor scenarios. Each dataset provides a sequence of 4-tuple of RGB image, depth map, wireless channel, and user location. Future work on the framework includes expanding the ViWi database of scenarios and incorporate more data processing features.
\label{conclusion}



\begin{thebibliography}{10}
	\providecommand{\url}[1]{#1}
	\csname url@samestyle\endcsname
	\providecommand{\newblock}{\relax}
	\providecommand{\bibinfo}[2]{#2}
	\providecommand{\BIBentrySTDinterwordspacing}{\spaceskip=0pt\relax}
	\providecommand{\BIBentryALTinterwordstretchfactor}{4}
	\providecommand{\BIBentryALTinterwordspacing}{\spaceskip=\fontdimen2\font plus
		\BIBentryALTinterwordstretchfactor\fontdimen3\font minus
		\fontdimen4\font\relax}
	\providecommand{\BIBforeignlanguage}[2]{{%
			\expandafter\ifx\csname l@#1\endcsname\relax
			\typeout{** WARNING: IEEEtran.bst: No hyphenation pattern has been}%
			\typeout{** loaded for the language `#1'. Using the pattern for}%
			\typeout{** the default language instead.}%
			\else
			\language=\csname l@#1\endcsname
			\fi
			#2}}
	\providecommand{\BIBdecl}{\relax}
	\BIBdecl
	
	\bibitem{Boccardi2014}
	F.~Boccardi, R.~Heath, A.~Lozano, T.~Marzetta, and P.~Popovski, ``Five
	disruptive technology directions for {5G},'' \emph{IEEE Communications
		Magazine}, vol.~52, no.~2, pp. 74--80, Feb. 2014.
	

	
	\bibitem{Rappaport2013a}
	T.~Rappaport, S.~Sun, R.~Mayzus, H.~Zhao, Y.~Azar, K.~Wang, G.~Wong, J.~Schulz,
	M.~Samimi, and F.~Gutierrez, ``Millimeter wave mobile communications for {5G}
	cellular: It will work!'' \emph{IEEE Access}, vol.~1, pp. 335--349, May 2013.
	
	\bibitem{Bjoernson2016}
	E.~Bjornson, E.~G. Larsson, and T.~L. Marzetta, ``Massive {MIMO}: Ten myths and
	one critical question,'' \emph{IEEE Communications Magazine}, vol.~54, no.~2,
	pp. 114--123, February 2016.
	
	\bibitem{Alkhateeb2018}
	A.~Alkhateeb, S.~Alex, P.~Varkey, Y.~Li, Q.~Qu, and D.~Tujkovic, ``Deep
	learning coordinated beamforming for highly-mobile millimeter wave systems,''
	\emph{IEEE Access}, vol.~6, pp. 37\,328--37\,348, 2018.
	
	\bibitem{LIS}
	A.~Taha, M.~Alrabeiah, and A.~Alkhateeb, ``Enabling large intelligent surfaces
	with compressive sensing and deep learning,'' \emph{arXiv preprint
		arXiv:1904.10136}, 2019.
	
	\bibitem{DLChMapping}
	M.~Alrabeiah and A.~Alkhateeb, ``Deep learning for {TDD} and {FDD} massive
	mimo: Mapping channels in space and frequency,'' \emph{arXiv preprint
		arXiv:1905.03761}, 2019.
	
	\bibitem{OShea2018}
	T.~J. {OShea}, T.~{Roy}, and T.~C. {Clancy}, ``Over-the-air deep learning based
	radio signal classification,'' \emph{IEEE Journal of Selected Topics in
		Signal Processing}, vol.~12, no.~1, pp. 168--179, Feb 2018.
	
	\bibitem{Alkhateeb2018a}
	A.~Alkhateeb, I.~Beltagy, and S.~Alex, ``Machine learning for reliable mmwave
	systems: Blockage prediction and proactive handoff,'' in \emph{IEEE
		GlobalSIP, arXiv preprint arXiv:1807.02723}, 2018.
	
	\bibitem{Li2019}
	X.~{Li} and A.~{Alkhateeb}, ``Deep learning for direct hybrid precoding in
	millimeter wave massive {MIMO} systems,'' \emph{in Proc. of Asilomar CSSC,
		arXiv e-prints}, p. arXiv:1905.13212, May 2019.
	
	\bibitem{Samuel2017}
	N.~{Samuel}, T.~{Diskin}, and A.~{Wiesel}, ``Deep mimo detection,'' in
	\emph{2017 IEEE 18th International Workshop on Signal Processing Advances in
		Wireless Communications (SPAWC)}, July 2017, pp. 1--5.
	
	\bibitem{Geiger2012}
	A.~Geiger, P.~Lenz, and R.~Urtasun, ``Are we ready for autonomous driving? the
	kitti vision benchmark suite,'' in \emph{Conference on Computer Vision and
		Pattern Recognition (CVPR)}, 2012.
	
	\bibitem{Russakovsky2015}
	O.~Russakovsky, J.~Deng, H.~Su, J.~Krause, S.~Satheesh, S.~Ma, Z.~Huang,
	A.~Karpathy, A.~Khosla, M.~Bernstein \emph{et~al.}, ``Imagenet large scale
	visual recognition challenge,'' \emph{International journal of computer
		vision}, vol. 115, no.~3, pp. 211--252, 2015.
	
	\bibitem{DeepMIMODataset}
	\BIBentryALTinterwordspacing
	\emph{DeepMIMO Dataset}. [Online]. Available: \url{https://www.DeepMIMO.net}
	\BIBentrySTDinterwordspacing
	
	\bibitem{Klautau}
	\BIBentryALTinterwordspacing
	A.~{Klautau}, N.~{González-Prelcic}, and R.~W. {Heath}, ``Lidar data for deep
	learning-based mmwave beam-selection,'' \emph{IEEE Wireless Communications
		Letters}, vol.~8, no.~3, pp. 909--912, June 2019. [Online]. Available:
	\url{https://www.lasse.ufpa.br/raymobtime/}
	\BIBentrySTDinterwordspacing
	
	
	
	\bibitem{ViWi}
	\BIBentryALTinterwordspacing
	\emph{ViWi Dataset}. [Online]. Available: \url{https://www.viwi-dataset.net}
	\BIBentrySTDinterwordspacing
	
	\bibitem{SynData-CV1}
	A.~Gaidon, Q.~Wang, Y.~Cabon, and E.~Vig, ``Virtual worlds as proxy for
	multi-object tracking analysis,'' in \emph{Proceedings of the IEEE conference
		on computer vision and pattern recognition}, 2016, pp. 4340--4349.
	
	\bibitem{Synthia}
	G.~Ros, L.~Sellart, J.~Materzynska, D.~Vazquez, and A.~M. Lopez, ``The synthia
	dataset: A large collection of synthetic images for semantic segmentation of
	urban scenes,'' in \emph{Proceedings of the IEEE conference on computer
		vision and pattern recognition}, 2016, pp. 3234--3243.
	
	\bibitem{SIM4CV}
	M.~M{\"u}ller, V.~Casser, J.~Lahoud, N.~Smith, and B.~Ghanem, ``Sim4cv: A
	photo-realistic simulator for computer vision applications,''
	\emph{International Journal of Computer Vision}, vol. 126, no.~9, pp.
	902--919, 2018.
	
	\bibitem{Alrabeiah2019a}
	M.~{Alrabeiah} and A.~{Alkhateeb}, ``Deep learning for mmwave beam and blockage
	prediction using {Sub-6GHz} channels,'' \emph{IEEE Transactions
		on Communications, arXiv e-prints}, p. arXiv:1910.02900, Oct 2019.
	
\end{thebibliography}
\end{document}